\documentclass[conference]{IEEEtran}
\IEEEoverridecommandlockouts

\usepackage{amsmath,amssymb,amsfonts}
\usepackage{algorithmic}
\usepackage{graphicx}
\usepackage{textcomp}
\usepackage{xcolor}
\usepackage{booktabs}
\usepackage{svg}
\usepackage{booktabs}
\usepackage{multirow}
\usepackage{siunitx}
\usepackage{float} % Required for [H] option
\usepackage{tikz}
\usepackage{subfigure}
\usepackage{pgfplots} % for creating plots with TikZ
\def\BibTeX{{\rm B\kern-.05em{\sc i\kern-.025em b}\kern-.08em
    T\kern-.1667em\lower.7ex\hbox{E}\kern-.125emX}}
\usepackage[breaklinks,colorlinks]{hyperref}

\usepackage[%
  xindy,
  acronym,
]{glossaries}
\glsdisablehyper

\usepackage[capitalize]{cleveref}

\newacronym{gl:LoD}{LoD}{Level of Detail}
\newacronym{gl:U-Net}{U-Net}{U-shape Neural Network}
\newacronym{gl:FCN}{FCN}{Fully Convolutional Network} 
\newacronym{gl:AFM}{AFM}{Attraction Field Map}
\newacronym{gl:GIS}{GIS}{Geographic Information System}
\newacronym{gl:nDSM}{nDSM}{normalized Digital Surface Model}
\newacronym{gl:PAN}{PAN}{panchromatic}
\newacronym{gl:GNN}{GNN}{Graph Neural Network}
\newacronym{gl:VHR}{VHR}{Very High Resolution}
\newacronym{gl:RMSE}{RMSE}{Root Mean Square Error}
\newacronym{gl:PCA}{PCA}{Principal Component Analysis}
\newacronym{gl:IoU}{IoU}{Intersection over Union}
\newacronym{gl:R2U-Net}{R2U-Net}{Residual Recurrent U-Net}
\newacronym{gl:R2AU-Net}{R2AU-Net}{Residual Recurrent Attention U-Net}

\newcommand{\RePolyWorlddot}{Re:PolyWorld }
\newcommand{\RePolyWorld}{\RePolyWorlddot\space}

\begin{document}

\title{PolyRoof: Precision Roof Polygonization in Urban Residential Building with Graph Neural Networks}
%PolyRoof: Polygonal Roof Extraction with Graph Neural Networks in Remote Sensing Images
% RoofGraph
%RoofGeometry: Polygonal Roof Geometry Extraction with Graph Neural Networks in Remote Sensing Images
% PolyRoof: Precision Roof Modeling in Urban Landscapes with Graph Neural Networks
% PolyRoof: Precision Roof Polygonization in Urban Landscapes with Graph Neural Networks

\author{
\IEEEauthorblockN{\fontsize{12}{12}\selectfont 
    1\textsuperscript{st} Chaikal Amrullah \hspace{2cm} 
    2\textsuperscript{nd} Daniel Panangian \hspace{2cm} 
    3\textsuperscript{rd} Ksenia Bittner}
\vspace{0.2cm}
\IEEEauthorblockA{
\parbox{1\textwidth}{
\centering
\textit{\fontsize{11}{11}\selectfont Remote Sensing Technology Institute (IMF)} \\
\vspace{0.1cm}
\textit{\fontsize{11}{11}\selectfont German Aerospace Center (DLR)}\\
\vspace{0.1cm}
\textit{\fontsize{11}{11}\selectfont Oberpfaffenhofen, Germany} \\
\vspace{0.1cm}
\textit{\fontsize{11}{11}\selectfont 
    chaikal.amrullah@dlr.de \hspace{0.5cm} 
    daniel.panangian@dlr.de \hspace{0.5cm} 
    ksenia.bittner@dlr.de}
}}}

\maketitle
\thispagestyle{plain}
\pagestyle{plain}

\begin{abstract}
The growing demand for detailed building roof data has driven the development of automated extraction methods to overcome the inefficiencies of traditional approaches, particularly in handling complex variations in building geometries. Re:PolyWorld, which integrates point detection with graph neural networks, presents a promising solution for reconstructing high-detail building roof vector data. This study enhances Re:PolyWorld’s performance on complex urban residential structures by incorporating attention-based backbones and additional area segmentation loss. Despite dataset limitations, our experiments demonstrated improvements in point position accuracy (1.33 pixels) and line distance accuracy (14.39 pixels), along with a notable increase in the reconstruction score to 91.99\%. These findings highlight the potential of advanced neural network architectures in addressing the challenges of complex urban residential geometries.
\end{abstract}


\glsresetall

\section{Introduction}

The demand for high-\gls{gl:LoD} building data has grown significantly, driven by the need to accurately capture complex geometric features \cite{kutzner2020citygml}. This data is essential for a wide range of applications, including urban development planning, architectural design, construction, and infrastructure monitoring \cite{bittner2018building, schuegraf2019automatic}. However, producing detailed and high-quality building data remains a significant challenge. Conventional methods, such as stereo-plotting from aerial imagery, are inefficient, labor-intensive, and time-consuming \cite{10261_19819}.

\begin{figure}[!tbp]
    \centering
    \begin{tikzpicture}
        % First Row of Subfigures
        \node (img1) at (0, 0) {\includegraphics[width=0.22\linewidth, trim=2cm 2cm 2cm 2cm, clip]{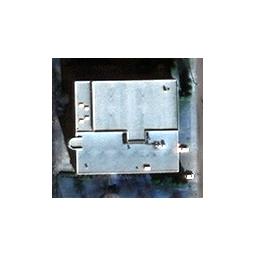}};
        \node (img2) at (2, 0) {\includegraphics[width=0.22\linewidth, trim=1.25cm 1.25cm 1.25cm 1.5cm, clip]{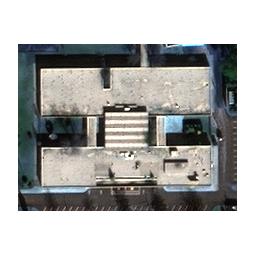}};
        \node (img3) at (4.5, 0) {\includegraphics[width=0.22\linewidth]{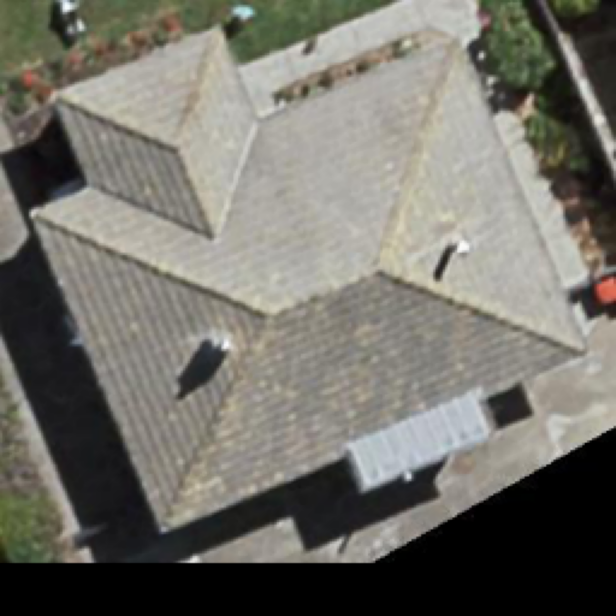}};
        \node (img4) at (6.5, 0) {\includegraphics[width=0.22\linewidth]{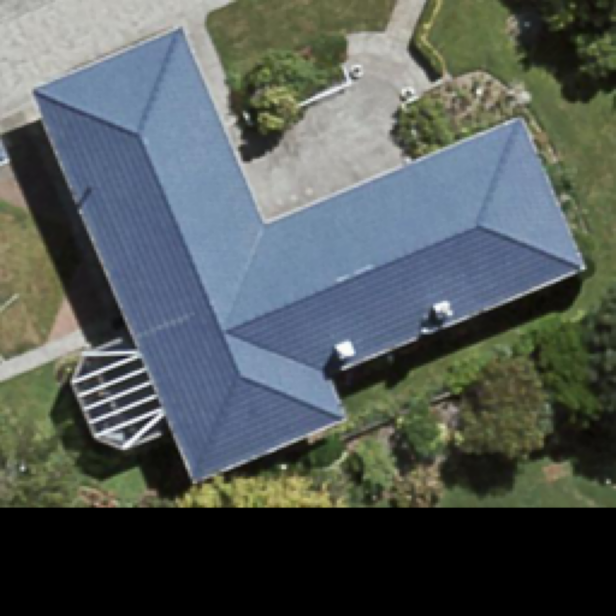}};
        
        % Second Row of Subfigures
        \node (img5) at (0, -2.5) {\includegraphics[width=0.22\linewidth, trim=2.3cm 2.3cm 2.3cm 2.3cm, clip]{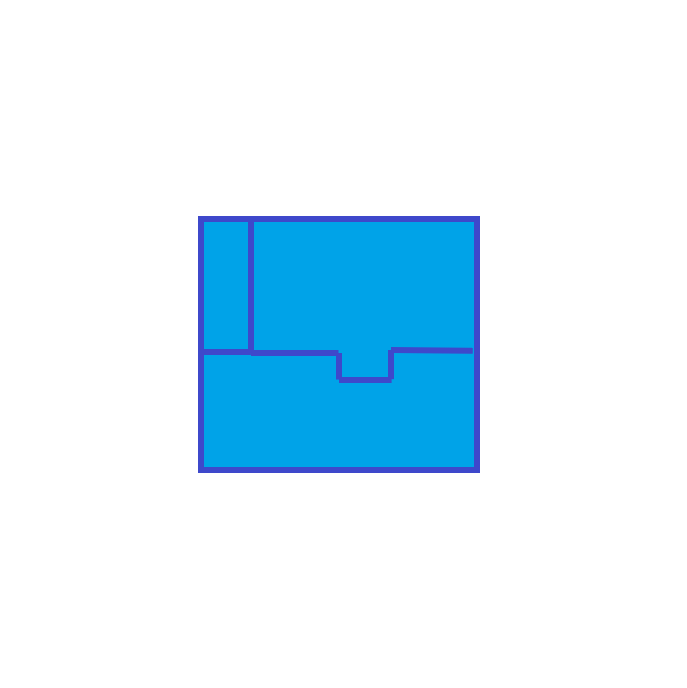}};
        \node (img6) at (2, -2.5) {\includegraphics[width=0.22\linewidth, trim=1.9cm 1.9cm 1.9cm 1.9cm, clip]{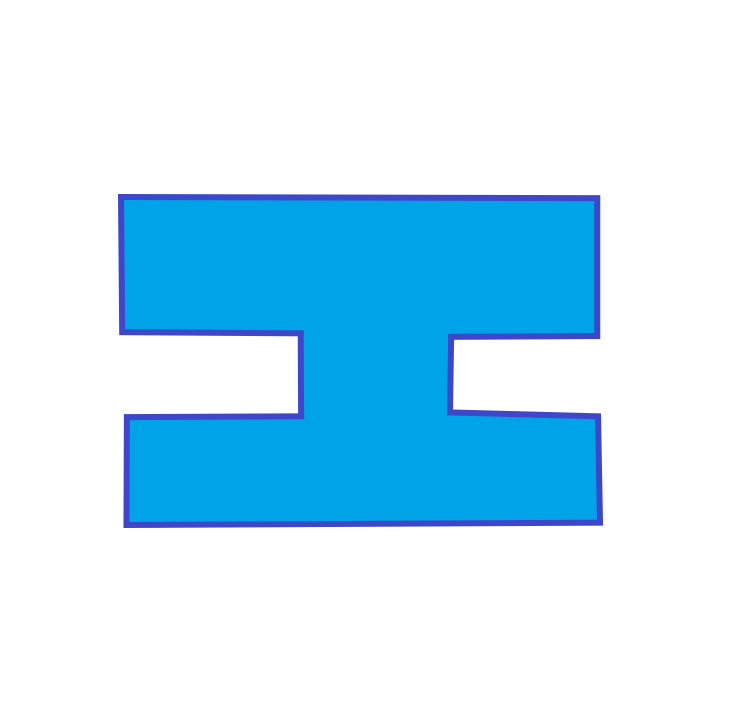}};
        \node (img7) at (4.5, -2.5) {\includegraphics[width=0.22\linewidth]{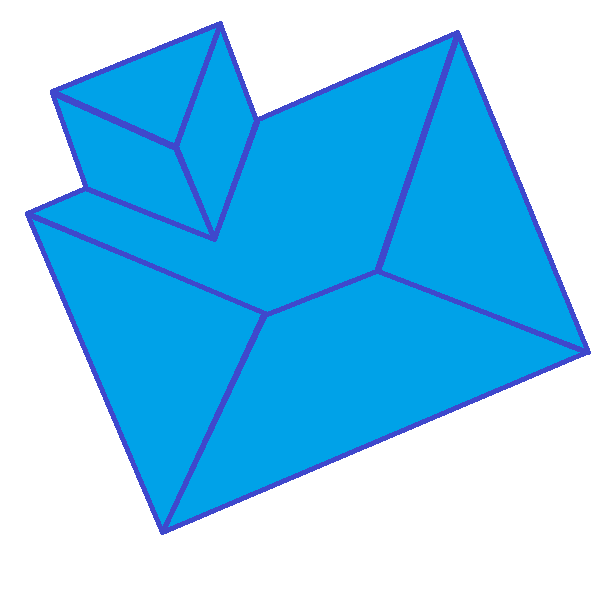}};
        \node (img8) at (6.5, -2.5) {\includegraphics[width=0.22\linewidth]{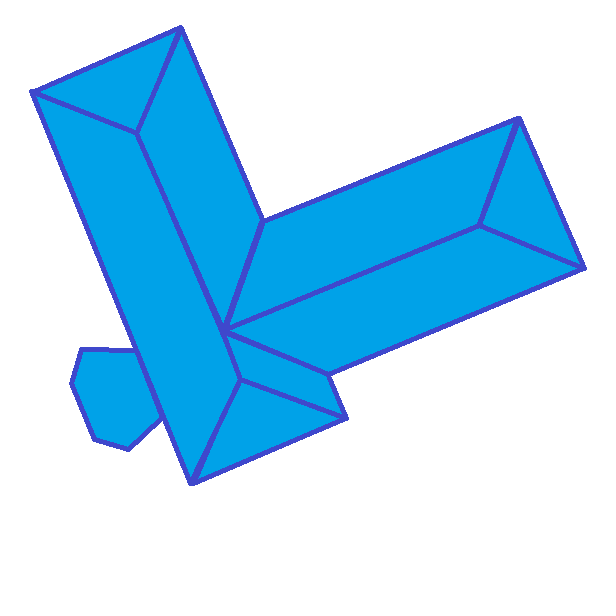}};
        
        % Vertical Line Overlay
        \draw[thick] (3.25, 0.8) -- (3.25, -3); % Adjust coordinates for positioning
    \end{tikzpicture}
    
    \caption{Difference of industrial (column one and two) and urban residential (column three and four) building complexity}
    \label{fig:industrial_urban}
\end{figure}

In recent years, automatic building data extraction methods, particularly for roof structures, have gained increasing attention across various fields \cite{bittner2018building, schuegraf2019automatic, 10261_19819, qian2022deep, zhao2022extracting, zorzi2023re}. Extracting roof structures involves capturing fine-grained variations in size, shape, and spatial distribution \cite{qian2022deep}, alongside complex features such as edges, corners, inscriptions, and roof components. A key challenge in this process lies not only in managing the diversity of object textures but also in addressing the inherent complexity of building geometries. Other challenges also arise when discussing the complexity variations that also change based on the type of construction activity, be it commercial, industrial, health or residential activities (see \Cref{fig:industrial_urban}). These geometric challenges include irregular shapes, varying roof angles, and overlapping components \cite{pantazis2019beyond}, all of which require precise modeling to achieve accurate and reliable reconstruction.

High-\gls{gl:LoD} building data has been pursued through various methods, with texture-based approaches such as semantic segmentation using \Glspl{gl:FCN} and \Glspl{gl:U-Net} architectures \cite{ronneberger2015u} improving boundary delineation and adjacent structure separation. Instance-level segmentation has advanced extraction capabilities but struggles in scenarios with densely populated areas, complex junctions, or internal courtyards \cite{he2018maskrcnn, schuegraf2019automatic}. As high-\gls{gl:LoD} building data is often represented as vector polygons, methods targeting the geometric complexity of these representations have gained prominence. Approaches like frame field-guided polygonization and \Glspl{gl:GNN} for modeling roof-lines \cite{zhao2022extracting} have proven effective. Notably, \gls{gl:GNN}-based frameworks \cite{zorzi2023re} excel in capturing spatial relationships and enabling detailed building reconstructions tailored to \gls{gl:GIS} applications.

In this study, we extend \RePolyWorld \cite{zorzi2023re}, which combines point detection with graph neural networks to address both the texture- and geometry-based complexities of building roof data extraction. By automatically detecting object vertices and solving an optimal transport problem, this model generates accurate polygonal representations of roof structures.

We propose \textsc{PolyRoof}, by implementing model architecture improvement, data augmentation and  tailored additional area segmentation loss to improve building reconstruction using datasets of urban residential complexities \cite{ren2021intuitiveefficientroofmodeling}. A major challenge is addressing the diverse geometric characteristics of different scenes, from simpler, repetitive industrial structures \cite{chen2022heatholisticedgeattention} to more complex and relatively irregular urban residential with densely packed roofs as shown in \Cref{fig:industrial_urban}. PolyRoof aims to tackle this challenge, advancing automated building data extraction for detail-oriented applications across varied urban landscapes.

\section{Dataset}
For this study, we used the RoofOpt dataset \cite{ren2021intuitiveefficientroofmodeling} for its \gls{gl:VHR} RGB aerial imagery and detailed annotations of residential roofs with complex geometries, including non-convex shapes and irregular angles. These challenges make it ideal for evaluating polygonization models. In contrast, the original \RePolyWorld utilized the HEAT dataset \cite{chen2022heatholisticedgeattention}, which includes industrial roof segment data called outdoor and floor plan data, both annotated with vector representations of edges while outdoor has RGB aerial image and floor plan has density image. \RePolyWorld demonstrated strong performance on both HEAT components.

Given the varying geometric complexities across different scenes, we compared our urban residential RoofOpt dataset with the HEAT dataset to better understand these differences. Key geometric attributes—number of vertices, point degree, convexity, compactness, and the second component of \gls{gl:PCA} (referred to here as the \gls{gl:PCA} score)—were analyzed to evaluate their impact on model performance. The number of vertices indicates polygon intricacy, while point degree reflects structural connectivity. Convexity and compactness measure shape regularity and circularity, offering insights into variability. The \gls{gl:PCA} score consolidates these attributes into a single orthogonal metric, highlighting geometric diversity across datasets.

This analysis aims to statistically quantify the differences in geometric complexity between the datasets used in this study and those used previously summarized in \cref{tab:geometry_statistic_dataset}.

\begin{table}[!h]
\centering
\caption{Average Building Geometry Complexity.}
\small % Reduce the font size of the table
\scalebox{0.85}{
\begin{tabular}{p{0.225\columnwidth} r r r r r} % 6 columns
\toprule
\textbf{Geometry}       & \textbf{Num.}     & \textbf{Point}      & \textbf{Conv-}      & \textbf{Compa-}      & \textbf{\gls{gl:PCA}} \\
\textbf{Property}       & \textbf{Vertices} $\uparrow$ & \textbf{Degree} $\uparrow$ & \textbf{exity} $\uparrow$ & \textbf{ctness} $\downarrow$ & \textbf{Score} $\uparrow$ \\
\midrule
RoofOpt \cite{ren2021intuitiveefficientroofmodeling}   & \textbf{36.32} & 6.09  & \textbf{90.26} & \textbf{11.56} & \textbf{42.83} \\
Outdoor \cite{chen2022heatholisticedgeattention}       & 12.55        & 5.42  & 87.06         & 61.07         & 29.21         \\
Floorplan \cite{chen2022heatholisticedgeattention}     & 21.95        & \textbf{6.97} & 86.17         & 65.29         & 38.80         \\
\bottomrule
\end{tabular}}
\label{tab:geometry_statistic_dataset}
\end{table}

The dataset analysis reveals significant geometric differences: RoofOpt, with the highest mean vertex count (36.32) and mean convexity (90.26) but lowest compactness (11.83), reflects complex roof geometries in relatively sparse area. In contrast, Outdoor data, with fewer vertices and higher compactness, represents simpler structures with low spread. \gls{gl:PCA} scores highlight RoofOpt's complexity (42.83), followed by Floorplan (38.80) and Outdoor (29.21), emphasizing the challenge of handling high-detail datasets like RoofOpt.

\begin{figure}
\centering
\includegraphics[width=0.8\linewidth]{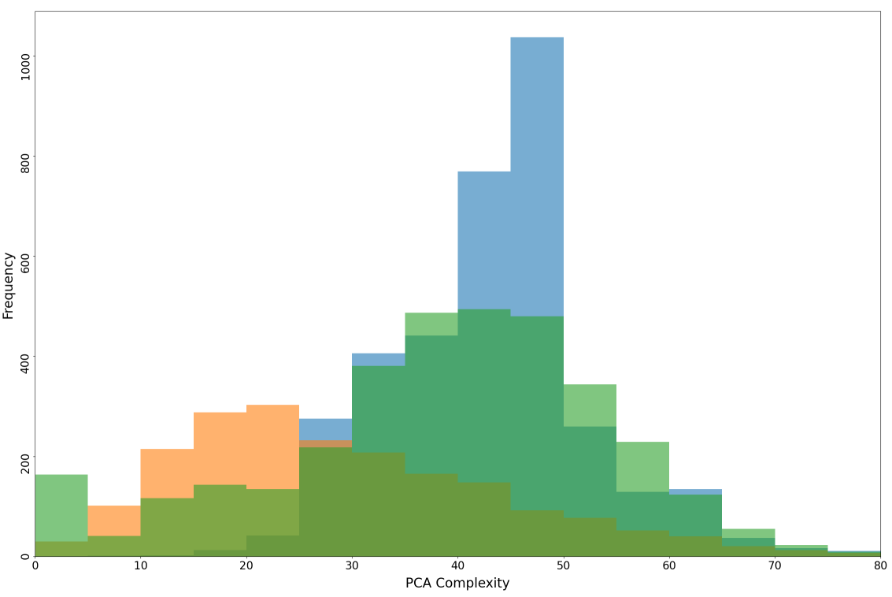}
\caption{Dataset histogram based on \gls{gl:PCA} complexity score of geometry complexity. RoofOpt \cite{ren2021intuitiveefficientroofmodeling} dataset with blue while HEAT \cite{chen2022heatholisticedgeattention} dataset for outdoor is in orange and floorplan with green.}
\label{fig:dataset_histogram}
\end{figure}

\Cref{fig:dataset_histogram} illustrates \gls{gl:PCA} score distributions. RoofOpt's histogram shows a medium-high concentration with high kurtosis, indicating uneven data distribution. In contrast, HEAT exhibits a near-normal distribution with presence of skewness, suggesting better uniformity. 

Using PCA complexity scores, a balanced split into training, validation, and test sets ensured representation of all complexity levels. This strategy enhances model generalization and performance across varying geometric complexities.\\

\section{Methodology}
From the dataset's geometric complexity analysis, we hypothesize that improving model enhancement strategies and evaluation metrics is crucial for urban residential building reconstruction. Enhancements include attention mechanisms and data augmentation for point detection, along with an additional loss function for matching-optimization, evaluated using metrics tailored to capture high-\gls{gl:LoD} intricacies.

\subsection{Model Enhancement Strategies}
\subsubsection{How data augmentation techniques and backbone architecture variations influence the generalization ability and precision of the PolyRoof Model?}

To enhance generalization in roof extraction models and address dataset size, variation, and distribution limitations, we employed extensive augmentation strategies, including rotations, flips, and adjustments to brightness, contrast, and sharpness, simulating real-world conditions. Additionally, we compared backbone architectures—\gls{gl:R2U-Net} \cite{alom2018recurrentresidualconvolutionalneural} and  \gls{gl:R2AU-Net} \cite{1360580628164331648}—to assess the impact of attention mechanisms in refining spatial precision. This analysis aimed to identify the optimal architecture balancing efficiency and segmentation accuracy.

\subsubsection{How additional loss functions enhance the model's performance in detail point detection and segmentation tasks?}

Optimizing the learning process with advanced loss functions is crucial for guiding and enhancing model performance. Traditional loss functions—such as vertex detection and matching, also angle and segmentation by option, losses—effectively evaluate corner point accuracy and relationships but struggle to address precise segmentation and the irregularities of complex roof geometries.

To address these gaps, we propose an additional loss functions: \textit{Area Segmentation Loss} ($\mathcal{L}{\text{$_{segm}$}}$). This value complements this by evaluating the F1-Score across three dimensions: building instances, roof segment instances, and a reconstruction index, which reflects the harmonic mean ratio of the building instances and the roof segment instances. This combined approach enhances both point accuracy and segmentation consistency, fostering improvements in the detailed reconstruction of building and roof structures.
\begin{equation}
    \mathcal{L}_{\text{$_{segm}$}} =  \text{F1}_{\text{building}} + \text{F1}_{\text{roof segment}} + \text{F1}_{\text{reconstruction}} 
    \label{eq:segmentation_area_loss}
\end{equation}

\subsection{Evaluation Metrics}
Recognizing the limitations of Average Precision (AP) and Average Recall (AR) scores in capturing the complexities of high-resolution data, we introduced alternative metrics to evaluate geometric primitives at multiple levels: Point Position Accuracy, Line Distance Accuracy, Building Instance F1-Score, Roof Segment F1-Score, and Reconstruction Score. These metrics, all within 50\% \gls{gl:IoU} threshold framework, provide a holistic assessment of prediction quality, aligning with the model's additional loss functions to evaluate accuracy, structural fidelity, and segmentation quality.

\paragraph{Point Position Accuracy}
This metric computes the RMSE between predicted and ground truth roof corner points, providing a measure of positional precision in pixel units.

\paragraph{Line Distance Accuracy} 
By averaging the Hausdorff and Fréchet distances, this metric captures both maximum deviation and sequential alignment, offering a comprehensive view of structural similarity.

\paragraph{Building Instance and Roof Segment F1-Scores} 
These metrics evaluate segmentation quality based on \gls{gl:IoU}, focusing on both the overall building shape (Building Instance F1) and individual roof segments (Roof Segment F1).

\paragraph{Reconstruction Score}
The harmonic mean of the Building Instance and Roof Segment F1-Scores, this score balances segmentation performance between full building shapes and their roof segments.
\\

\section{Result and Discussion}

\Cref{tab:experiment_metrics} presents the quantitative evaluation of each experimental configuration using the defined metrics, with \RePolyWorlddot serving as the Baseline for improvement variations. \Cref{fig:comparison-3d} illustrates qualitative prediction samples, anticipate a challenge linked to dataset geometry complexity in achieving high accuracy. Roof segment color indicates segmentation ID.

\begin{table}[!h]
\centering
\caption{Results of different experiment configurations}
\small % Reduce the font size of the table
\scalebox{0.85}{
\begin{tabular}{p{0.2\columnwidth} r r r r r} % 6 columns
\toprule
\textbf{} & \textbf{Point} & \textbf{Line} & \textbf{Building} & \textbf{Roof} & \textbf{Recon} \\
\textbf{Configuration} & \textbf{Pos. Acc.} & \textbf{Dist. Acc.} & \textbf{F1-Score} & \textbf{F1-Score} & \textbf{Score} \\
 & Pixel(s) $\downarrow$ & Pixel(s) $\downarrow$ & (\%) $\uparrow$ & (\%) $\uparrow$ & (\%) $\uparrow$ \\
\midrule
Baseline (B) & 1.34 & 18.31 & 89.38 & \textbf{89.68} & 88.57 \\
\midrule
B +Att & 1.26 & 17.01 & 88.36 & 89.06 & 87.75 \\
B +Aug & 1.34 & 20.72 & 62.28 & 88.39 & 72.11 \\
B +$\mathcal{L}_{\text{segm}}$ & 1.35 & 16.79 & 90.05 & 89.32 & 88.72 \\
\midrule
B +Att & \textbf{1.33} & \textbf{14.39} & \textbf{96.61} & 89.55 & \textbf{91.99} \\
\hspace{6 pt} +$\mathcal{L}_{\text{segm}}$ &  &  &  &  &  \\
\bottomrule
\end{tabular}
}
\vspace{0.25em}
\label{tab:experiment_metrics}
\end{table}

\newcolumntype{M}[1]{>{\centering\arraybackslash}m{#1}}
\begin{figure*}[!t]
    \centering
    \begin{tabular}{M{23mm}M{23mm}M{23mm}M{23mm}M{23mm}M{23mm}M{23mm}} % Adjust width for six columns
        \toprule
        Input & B & B +Att. & B +Aug. & B +$\mathcal{L}_{\text{segm}}$ & B +Att. +$\mathcal{L}_{\text{segm}}$ \\ % Add headers for six columns
        % Input & B & B +Att. & B +Aug. & B +$\mathcal{L}_{\text{segm}}$ & B +Att +$\mathcal{L}_{\text{segm}}$\\ % Add headers for six columns
        \midrule
        
        \includegraphics[width=\linewidth]{test_area_seg/CG10_500_041072_0050_gt.png} 
        & \includegraphics[width=\linewidth]{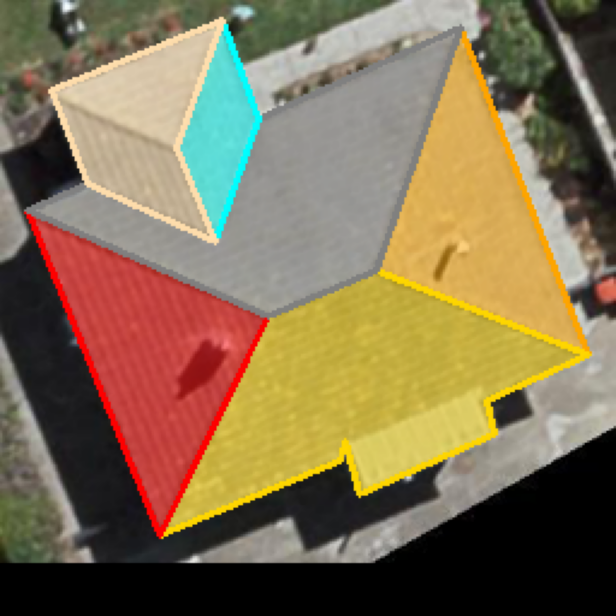} 
        & \includegraphics[width=\linewidth]{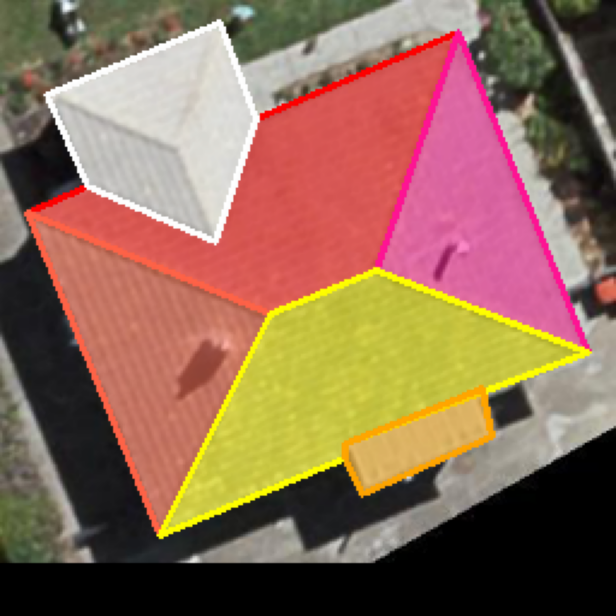}
        & \includegraphics[width=\linewidth]{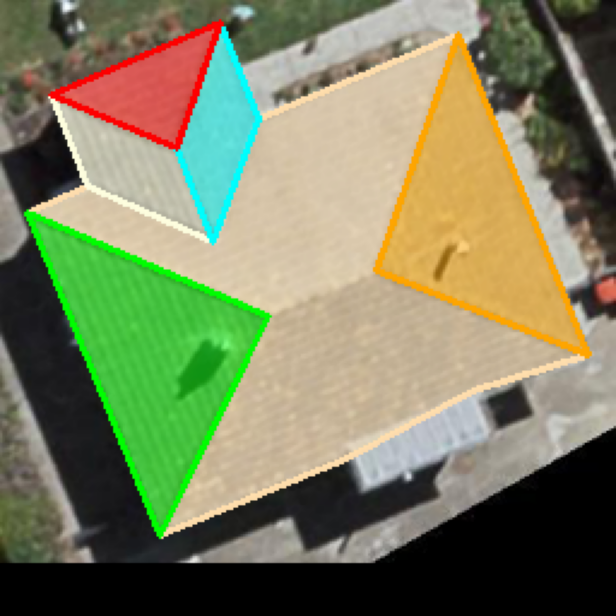} 
        & \includegraphics[width=\linewidth]{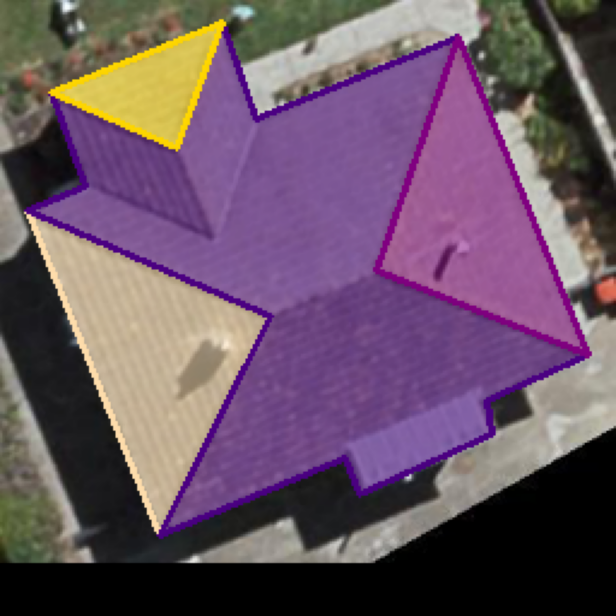} 
        & \includegraphics[width=\linewidth]{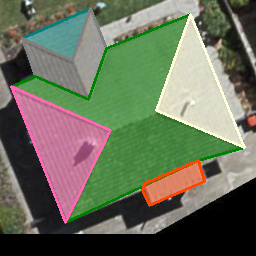} \\
        
        \includegraphics[width=\linewidth]{test_area_seg/CG10_500_045072_0054_gt.png}
        & \includegraphics[width=\linewidth]{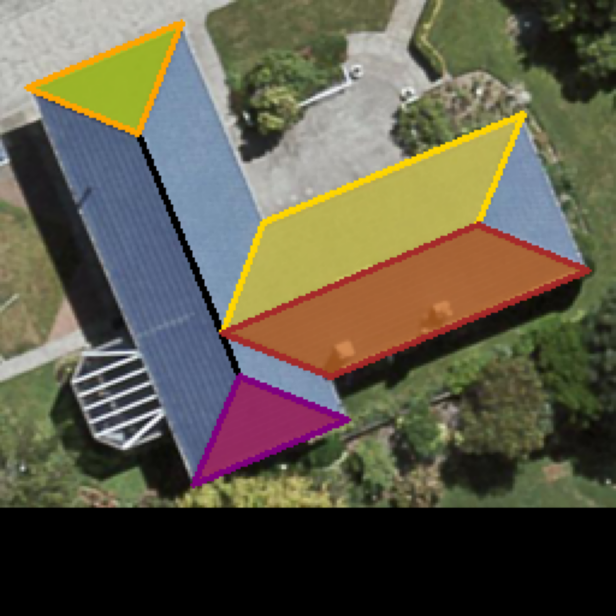} 
        & \includegraphics[width=\linewidth]{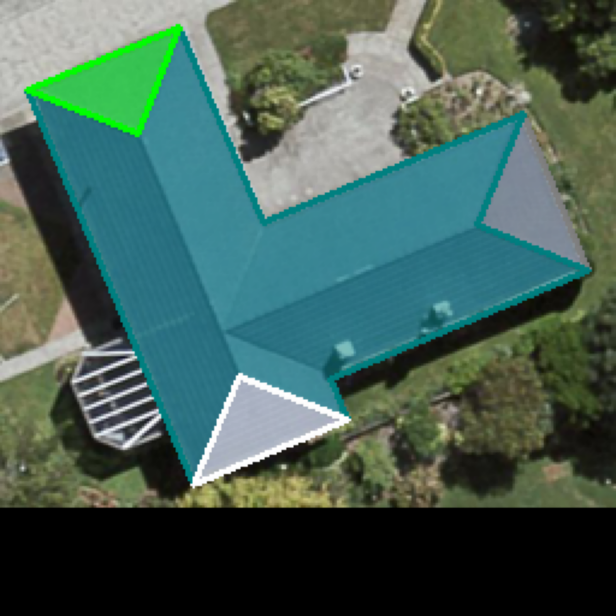}
        & \includegraphics[width=\linewidth]{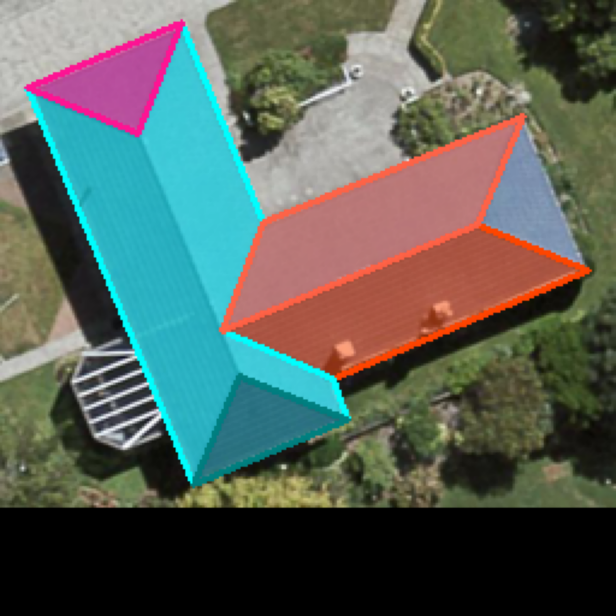} 
        & \includegraphics[width=\linewidth]{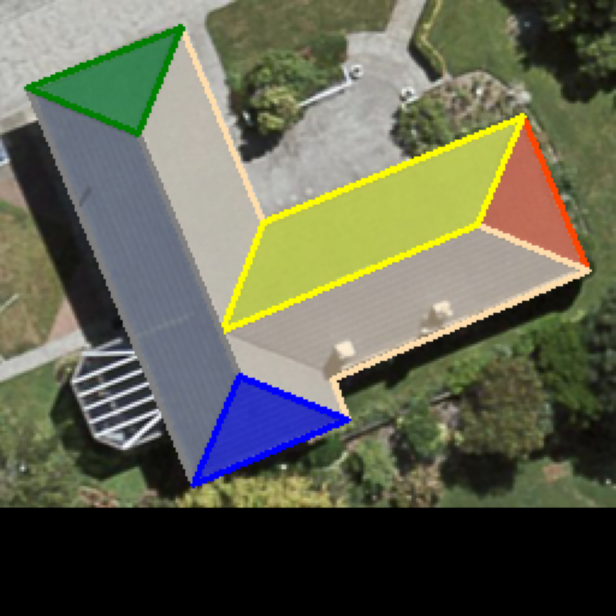}
        & \includegraphics[width=\linewidth]{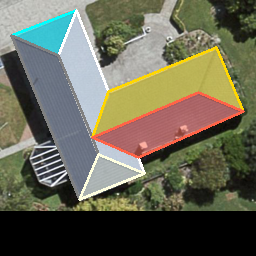}\\
        \bottomrule
    \end{tabular}
    \centering
    \caption{Visual comparison of RGB with various experiment scenario. All examples are taken from the test set.}
    \label{fig:comparison-3d}
\end{figure*}

The Baseline model showed balanced performance across our evaluation metrics, but its results were lower compared to those reported in the original paper \cite{zorzi2023re}, which used a dataset with lower geometric complexity \cite{chen2022heatholisticedgeattention} (Figure \ref{fig:dataset_histogram}). The original paper’s The \( F1\text{-}Score_{50} \), was 91.89\%, while the Building F1-Score in this paper was 89.38\%. Although Roof Segment performance remained similar, the model produced incomplete building reconstructions, leading to a lower Reconstruction Score of 88.57\%. This highlights the difficulty of reconstructing entire buildings with high accuracy on more complex datasets.

Adding attention mechanisms (+Att) improved point position accuracy, initially producing the best RMSE of 1.26 pixels. In the same time, incorporating $\mathcal{L}{\text{$_{segm}$}}$ improved Line Distance Accuracy to 16.79 pixels. When both techniques were combined, the result improved to 14.39 pixels. However, dataset augmentation led to a decrease in overall performance, suggesting caution when applying augmentation in point detection pipelines. On the other hand, $\mathcal{L}{\text{$_{segm}$}}$ showed the best Building Segmentation Score (90.05\%) and competitive Roof F1-Score (89.32\%), which directly impacted the Reconstruction Score (88.72\%).

When combining +Att with $\mathcal{L}_{\text{$_{segm}$}}$, we observed clear improvements across metrics, with Point Position Accuracy improving to 1.33 pixels, Line Distance Accuracy to 14.39 pixels, Building Instance F1-Score to 96.61\%, and the Reconstruction Score rising to 91.99\%. These were the best results overall, except for Roof F1-Score (89.55\%), which ranked second compared to the Baseline with fraction of gap. The combined configuration clearly boosted position accuracy and segmentation precision, resulting in a better overall score.

Qualitatively, while the combined approach outperformed the Baseline in general segmentation precision score, some roof segment edges were missing, leading to merged segments and duplicated roof segment IDs indicating completeness issues. Additionally, smaller or less common roof components were occasionally missed, indicating some consistency issues. These challenges highlight areas that require further refinement.

\glsresetall

\section{Conclusion}
The complexity of the dataset, with intricate roof structures and diverse shapes, significantly impacts model performance. While the model demonstrates potential for high \gls{gl:LoD} building predictions, challenges remain in handling complex geometries. 

Quantitatively, the combination of attention mechanisms and area segmentation loss achieved a notable reconstruction score of 91.99\%. Qualitatively, the model reconstructs buildings with detail and accuracy, though inconsistencies persist. These results underscore the need for further refinement, considering geometric complexity to enhance generalization and reliability in high \gls{gl:LoD} building model reconstruction.

\section*{Acknowledgment}
Chaikal Amrullah is currently funded by a DLR-DAAD Research Fellowship (No. 57681552) to pursue his PhD studies.

%\printbibliography
\bibliography{bibliography}
\end{document}